\documentclass[sigconf]{acmart}
\usepackage{enumitem}
\AtBeginDocument{%
  \providecommand\BibTeX{{%
    \normalfont B\kern-0.5em{\scshape i\kern-0.25em b}\kern-0.8em\TeX}}}

\copyrightyear{2023}
\acmYear{2023}
\setcopyright{rightsretained}
\acmConference[MM '23]{Proceedings of the 31st ACM International Conference on Multimedia}{October 29-November 3, 2023}{Ottawa, ON, Canada}
\acmBooktitle{Proceedings of the 31st ACM International Conference on Multimedia (MM '23), October 29-November 3, 2023, Ottawa, ON, Canada}
\acmDOI{10.1145/3581783.3612498}
\acmISBN{979-8-4007-0108-5/23/10}

\usepackage{multirow}
\usepackage{bm}

\begin{document}

\fancyhead{} 
\title{Pro-Cap: Leveraging a Frozen Vision-Language Model\\for Hateful Meme Detection}

\author{Rui Cao}
\email{ruicao.2020@phdcs.smu.edu.sg}
\affiliation{%
  \institution{Singapore Management University}
  \country{Singapore}
  \city{Singapore}
 }

\author{Ming Shan Hee}
\email{mingshan\_hee@mymail.sutd.edu.sg}
\affiliation{%
  \institution{Singapore University of Design and Technology}
  \country{Singapore}
  \city{Singapore}
 }

\author{Adriel Kuek}
\email{adrielkuek@gmail.com}
\affiliation{%
  \institution{DSO National Laboratories}
  \country{Singapore}
  \city{Singapore}
 }

\author{Wen-Haw Chong}
\email{whchong.2013@phdis.smu.edu.sg}
\affiliation{%
  \institution{Singapore Management University}
  \country{Singapore}
  \city{Singapore}
 }

\author{Roy Ka-Wei Lee}
\email{roy\_lee@sutd.edu.sg}
\affiliation{%
  \institution{Singapore University of Design and Technology}
  \country{Singapore}
  \city{Singapore}
 }

\author{Jing Jiang}
\email{jingjiang@smu.edu.sg}
\affiliation{%
  \institution{Singapore Management University}
  \country{Singapore}
  \city{Singapore}
 }

\renewcommand{\shortauthors}{Cao and Mingshan, et al.}
\begin{abstract}
Hateful meme detection is a challenging multimodal task that requires comprehension of both vision and language, as well as cross-modal interactions. Recent studies have tried to fine-tune pre-trained vision-language models (PVLMs) for this task. However, with increasing model sizes, it becomes important to leverage powerful PVLMs more efficiently, rather than simply fine-tuning them. Recently, researchers have attempted to convert meme images into textual captions and prompt language models for predictions. This approach has shown good performance but suffers from non-informative image captions. Considering the two factors mentioned above, we propose a probing-based captioning approach to leverage PVLMs in a zero-shot visual question answering (VQA) manner. Specifically, we prompt a frozen PVLM by asking hateful content-related questions and use the answers as image captions (which we call Pro-Cap), so that the captions contain information critical for hateful content detection. The good performance of models with Pro-Cap on three benchmarks validates the effectiveness and generalization of the proposed method.\footnote{Code is available at: https://github.com/Social-AI-Studio/Pro-Cap}
\end{abstract}




\maketitle 

{\color{red} \textbf{Disclaimer}: \textit{
This paper contains violence and discriminatory content that may be disturbing to some readers. 
}} 

\section{Introduction}
\label{sec:intro}
\begin{figure}[t] 
	
	\includegraphics[width=0.81\linewidth]{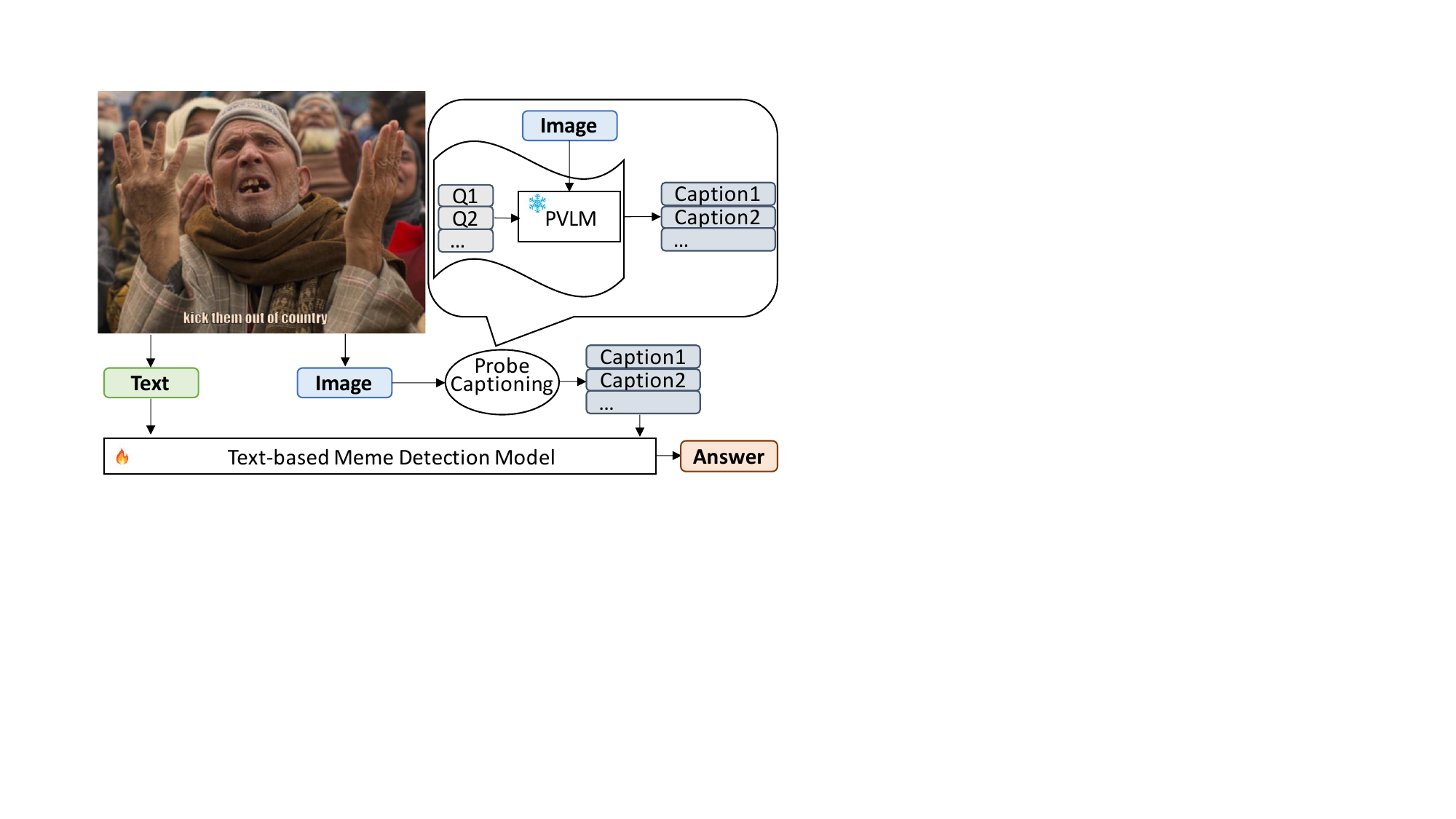} 
 \centering
\caption{The proposed probe-captioning approach. We prompt frozen pre-trained vision-language models via visual question answering to generate hateful content centric image captions.} 
	\label{fig:intro}
\end{figure}

Memes, which combine images with short texts, are a popular form of communication in online social media. Internet memes are often intended to express humor or satire. However, they are increasingly being exploited to spread hateful content across online platforms. Hateful memes attack individuals or communities based on their identities such as race, gender, or religion~\cite{DBLP:conf/nips/KielaFMGSRT20,DBLP:conf/acl/PramanickDMSANC21,DBLP:conf/semeval/FersiniGRSCRLS22,hee2023decoding}. The propagation of hateful memes can lead to discord online and may potentially result in hate crimes. Therefore, it is urgent to develop accurate hateful meme detection methods.

The task of hateful meme detection is challenging due to the multimodal nature of memes. Detection involves not only comprehending both the images and the texts but also understanding how these two modalities interact. Previous work~\cite{DBLP:conf/mm/LeeCFJC21,zhou2020multimodal,DBLP:conf/emnlp/PramanickSDAN021,zhu2022multimodal} learns cross-modal interactions from scratch using hateful meme detection datasets. However, it may be difficult for models to learn complicated multimodal interactions with the limited amount of data available from these datasets. With the development of Pre-trained Vision-Language Models (PVLMs) such as VisualBERT~\cite{li2019visualbert} and ViLBERT~\cite{lu2019vilbert}, recent work leverage these powerful PVLMs to facilitate the hateful meme detection task. A common approach is to fine-tune PVLMs with task-specific data~\cite{lippe2020multimodal,zhu2020enhance,DBLP:journals/corr/abs-2012-07788,DBLP:journals/corr/abs-2012-12975,hee2022explaining}. However, it is less feasible to fine-tune the larger models such as BLIP-2~\cite{DBLP:journals/corr/abs-2301-12597} and Flamingo~\cite{DBLP:journals/corr/abs-2204-14198} on meme detection because there are billions of trainable parameters. Therefore, computationally feasible solutions other than direct fine-tuning are needed to leverage large PVLMs in facilitating hateful meme detection.

Different from the approach above using PVLMs, \textsf{PromptHate}\cite{DBLP:conf/emnlp/CaoLC022} is a recently proposed model that converts the multimodal meme detection task into a unimodal masked language modeling task. It first generates meme image captions with an off-the-shelf image caption generator, ClipCap\cite{DBLP:journals/corr/abs-2111-09734}. By converting all input information into text, it can prompt a pre-trained language model along with two demonstrative examples to predict whether or not the input is hateful by leveraging the rich background knowledge in the language model. Although \textsf{PromptHate} achieves state-of-the-art performance, it is significantly affected by the quality of image captions, as shown in Table~\ref{tab:cap-impact}. Image captions that are merely generic descriptions of images may omit crucial details~\cite{DBLP:conf/mm/LeeCFJC21,zhu2020enhance}, such as the race and gender of people, which are essential for hateful content detection. But with additional image tags, such as entities found in the images and demographic information about the people in the images, the same model can be significantly improved, as shown in Table~\ref{tab:cap-impact}. However, generating these additional image tags is laborious and costly. For instance, entity extraction is usually conducted with the Google Vision Web Entity Detection API~\footnote{https://cloud.google.com/vision/docs/detecting-web}, which is a paid service. Ideally, we would like to find a more affordable way to obtain entity and demographic information from the images that is critical for hateful content detection.

\begin{table}[t]
\centering
\caption{Impact on detection performances on the FHM dataset~\cite{DBLP:conf/nips/KielaFMGSRT20} from image captions. 
  (w/o) denotes models without additional entity and demographic information.}
  \label{tab:cap-impact}
  \begin{tabular}{c|cc}
    \hline
    \textbf{Model}& \multicolumn{2}{c}{\textbf{Performance}}\\
   & AUC & Acc.\\
    \hline\hline
    PromptHate (w/o) & 76.76 & 67.28 \\
    PromptHate & 81.45 & 72.98 \\
    \hline
    VisualBERT (w/o) & 68.71&61.48 \\
    VisualBERT &72.56 &68.24\\
    ViLBERT (w/o) & 73.05&64.70\\
    ViLBERT & 75.72& 68.24\\
    \hline
\end{tabular}
\end{table}

Both above-mentioned approaches (i.e., one using PVLMs and the other converting the task to a unimodal task) have their pros and cons. In this paper, we combine the ideas from these two approaches and design a hateful meme detection method that leverages the power of a frozen PVLM to complement the unimodal approach of PromptHate. Specifically, we use a set of ``probing'' questions to query a PVLM (BLIP-2~\cite{DBLP:journals/corr/abs-2301-12597} in our experiments) for information related to common vulnerable targets in hateful content. The answers obtained from the probing questions will be treated as image captions (denoted as \textbf{Pro-Cap}) and used as input to a trainable hateful meme detection model. Figure~\ref{fig:intro} illustrates the overall workflow of the method. We refer to the step of using probing questions to generate the captions as \textit{probing-based captioning}.

Our proposed method fills existing research gaps by: 1) Leverage a PVLM without any adaptation or fine-tuning, thereby reducing computational cost; 2) Instead of explicitly obtaining additional image tags with costly APIs, we utilize the frozen PVLM to generate captions that contain information useful for hateful meme detection. To the best of our knowledge, this is the first work that to leverage PVLMs in a zero-shot manner through question answering to assist in the hateful meme detection task. To further validate our method, we test the effect of the generated Pro-Cap on both \textsf{PromptHate}\cite{DBLP:conf/emnlp/CaoLC022} and a BERT-based\cite{devlin2018bert} hateful meme detection model.


Based on the experimental results, we observe that \textsf{PromptHate} with Pro-Cap (denoted as Pro-Cap$\text{PromptHate}$) significantly surpasses the original \textsf{PromptHate} without additional image tags (i.e., about 4, 6, and 3 percentage points of absolute performance improvement on FHM~\cite{DBLP:conf/nips/KielaFMGSRT20}, MAMI~\cite{DBLP:conf/semeval/FersiniGRSCRLS22}, and HarM~\cite{DBLP:conf/emnlp/PramanickSDAN021} respectively). Pro-Cap$\text{PromptHate}$ also achieves comparable results with \textsf{PromptHate} with additional image tags, indicating that probing-based captioning can be a more affordable way of obtaining image entities or demographic information. Case studies further show that Pro-Cap offers essential image details for hateful content detection, enhancing the explainability of models to some extent. Meanwhile, Pro-Cap$_\text{BERT}$ clearly surpasses multimodal BERT-based models of similar sizes (i.e., about 7 percentage points of absolute improvement with VisualBERT on FHM~\cite{DBLP:conf/nips/KielaFMGSRT20}), proving the generalization of the probing-based captioning method.

\section{Related Work}
\label{sec:related}

\textit{Memes}, typically intended to be humorous or sarcastic, are increasingly being exploited for the proliferation of hateful content, leading to the challenging task of online hateful meme detection~\cite{DBLP:conf/semeval/FersiniGRSCRLS22,DBLP:conf/nips/KielaFMGSRT20,DBLP:conf/acl/PramanickDMSANC21}. To combat the spread of hateful memes, one line of work regards the hateful meme detection as a multimodal classification task. Researchers have applied pre-trained vision-language models (PVLMs) and fine-tune them based on meme detection data~\cite{zhu2020enhance,DBLP:journals/corr/abs-2012-12975,DBLP:journals/corr/abs-2012-07788,lippe2020multimodal}. To improve performance, some have tried model ensembling~\cite{DBLP:journals/corr/abs-2012-12975,DBLP:journals/corr/abs-2012-07788,lippe2020multimodal}. Another line of work considers combining pre-trained models (e.g., BERT~\cite{devlin2018bert} and CLIP~\cite{DBLP:conf/icml/RadfordKHRGASAM21}) with task-specific model architectures and tunes them end-to-end~\cite{DBLP:conf/emnlp/PramanickSDAN021,DBLP:conf/mm/LeeCFJC21,DBLP:journals/corr/abs-2210-05916}. Recently, authors in~\cite{DBLP:conf/emnlp/CaoLC022} have tried converting all meme information into text and prompting language models to better leverage the contextual background knowledge present in language models. This approach achieves the state-of-the-art results on two hateful meme detection benchmarks. However, it adopts a generic method for describing the image through image captioning, often ignoring important factors necessary for hateful meme detection. In this work, we seek to address this issue through probe-based captioning by prompting pre-trained vision-language models with hateful content-centric questions in a zero-shot VQA manner.

\section{Preliminary}
\label{sec:prelim}

We formally define our task and briefly review the use of pre-trained vision-language models (PVLMs) for zero-shot visual question answering (VQA). At the end of the section, we provide a brief introduction to the specific PVLM utilized in our work.

Given a meme image \( \mathcal{I} \) and a piece of accompanying meme text \( \mathcal{T} \), the model predicts whether the meme is hateful or not. Specifically, the model predicts scores \( \mathbf{s} \in \mathbb{R}^2 \) over the label space, where \( s_0 \) is a score indicating how likely the meme is \textit{non-hateful}, whereas \( s_1 \) is a score for the meme being \textit{hateful}. If \( s_0 > s_1 \), the model classifies the meme as non-hateful; otherwise, the meme is classified as hateful. Our proposed method (to be presented in detail in Section~\ref{sec:model}) uses zero-shot VQA to generate relevant captions to assist with hateful meme detection. To perform zero-shot VQA, we assume that there is a PVLM capable of processing an image and a textual prompt formatted as \textit{Question: }\texttt{[QUESTION]} \textit{Answer:}, where \texttt{[QUESTION]} is a placeholder for the question. The PVLM then generates a sequence of tokens as the answer to the question.
For example, given an image showing an Asian woman and the prompt \textit{Question: What is the race of the person in the image? Answer:}, the PVLM may generate the answer \textit{Asian}.

In this work, we use the recently released BLIP-2 model~\cite{DBLP:journals/corr/abs-2301-12597} as the PVLM, as it has demonstrated good performance in zero-shot VQA. The BLIP-2 model is composed of a frozen pre-trained image encoder, a frozen pre-trained language model, and a lightweight Querying Transformer, which is responsible for bridging the modality gap. It is worth noting that the BLIP-2 model can be replaced with any other PVLM that is capable of zero-shot VQA.


\section{Proposed Method}
\label{sec:model}


\subsection{Overview}

Recall that the key idea of our method is to elicit image details that are critical for hateful content detection, such as the gender and race of the people in the image. Because these details are not always included in automatically generated image captions, we propose relying on VQA to obtain such critical information, where the questions are carefully curated to elicit demographic and other relevant information. We opt to use zero-shot VQA because (1) for the intended type of questions, we do not have any VQA training data to train our own model, and (2) recent work has demonstrated promising performance of zero-shot VQA.

Specifically, we prompt the PVLM with $K$  \textit{probing questions} and regard the set of $K$ answers from the PVLM as image captions, which we refer to as \textbf{Pro-Cap}. We then combine the original text $\mathcal{T}$ with Pro-Cap as input to a hateful meme detection model. We experiment with two alternative hateful meme detection models: one based on BERT encoding, and the other based on PromptHate, a recently proposed prompting-based hateful meme detection model.

In the rest of this section, we first present the details of how we design our VQA questions to elicit the most critical details of an image for hateful meme detection. We then explain how the generated Pro-Cap is used by two alternative hateful meme detection models.


\subsection{Design of VQA Questions}

We leverage PVLMs for zero-shot VQA to generate Pro-Cap as image captions. We want Pro-Cap to provide not only a general description of the image but also details critical for hateful meme detection. To obtain a general caption of the image, we design the first probing question to inquire about the generic content of the image, as shown in Table~\ref{tab:prompt-ques}. However, such generic captions may be insufficient for hateful meme detection as hateful content usually targets persons or groups with specific characteristics, such as race, gender, or religion~\cite{DBLP:conf/semeval/FersiniGRSCRLS22,DBLP:conf/nips/KielaFMGSRT20}. Additionally, previous studies have shown that augmenting image representations with entities found in the image or demographic information of people in the image significantly aids hateful meme detection~\cite{zhu2020enhance,DBLP:conf/mm/LeeCFJC21}. Such details may be missing in generic image captions. Therefore, we design additional questions that aim to bring out information central to hateful content. This aligns the generated image captions more closely with the goal of hateful meme detection. Specifically, the high-level idea is to ask questions about common vulnerable targets of hateful content. Inspired by~\cite{mathias-etal-2021-findings}, which categorizes the targets of hateful memes into \textit{Religion}, \textit{Race}, \textit{Gender}, \textit{Nationality}, and \textit{Disability}, we ask questions about these five types of targets. For example, to generate image captions that indicate the race of the people in an image, we can ask the following question: \textit{what is the race of persons in the image?} We list the five questions designed for these five types of targets in Table~\ref{tab:prompt-ques}. Additionally, we observe that some animals, such as pigs, are often depicted in hateful memes, frequently as a means to annoy Muslims. With this consideration, we also design a question asking about the presence of animals in the image.


In~\cite{DBLP:journals/corr/abs-2210-07688}, the author claimed that PVLMs may hallucinate non-existent objects. For example, even when there is nobody in an image, PVLMs may generate an answer about race in response to the question \textit{what is the race of the person in the image?}. To prevent such misleading information, we use two validation questions. Specifically, we inquire about the existence of persons and animals. Only when the PVLM responds that a person or an animal exists will we include in the Pro-Cap the answers to those person-related or animal-related questions. For instance, if the answer to the question validating the existence of people indicates that nobody is present, we will ignore all answers from questions asking about \textit{religion}, \textit{race}, \textit{gender}, \textit{nationality}, and \textit{disability}.

We use $\mathcal{C}$ to represent the concatenation of the answers to the probing questions that are finally included as part of the Pro-Cap based on the validation results.
We will then concatenate $\mathcal{T}$ and $\mathcal{C}$ together as input to a purely text-based hateful meme classification model, as shown at the bottom of Figure~\ref{fig:intro}.
\begin{table}[ht]
\centering
\caption{Details of questions prompting PVLMs. The first block of the question asks about the content of the image; questions in the second block ask about commonly seen vulnerable targets in hateful contents; the last block questions validate the existence of persons and animals.}
\label{tab:prompt-ques}
  \begin{tabular}{l|p{4cm}} 
    \toprule
    \textbf{Focus} &  \textbf{Questions}\\
    \midrule
    \textbf{Content} & what is shown in the image?\\
    \midrule
    \textbf{Race} & What is the race of the person in the image?\\
    \textbf{Gender} & What is the gender of the person in the image? \\
    \textbf{Religion} & What is the religion of the person in the image?\\
    \textbf{Nationality} & Which country does the person in the image come from?\\
    \textbf{Disability} & Are there disabled people in the image?\\
    \textbf{Animal} & What animal is in the image?\\
    \midrule
    \textbf{Val Person} & Is there a person in the image?\\
    \textbf{Val Animal} & Is there an animal in the image?\\
    \bottomrule
\end{tabular}
\end{table}

\begin{figure}[t] 
	\includegraphics[width=0.85\linewidth]{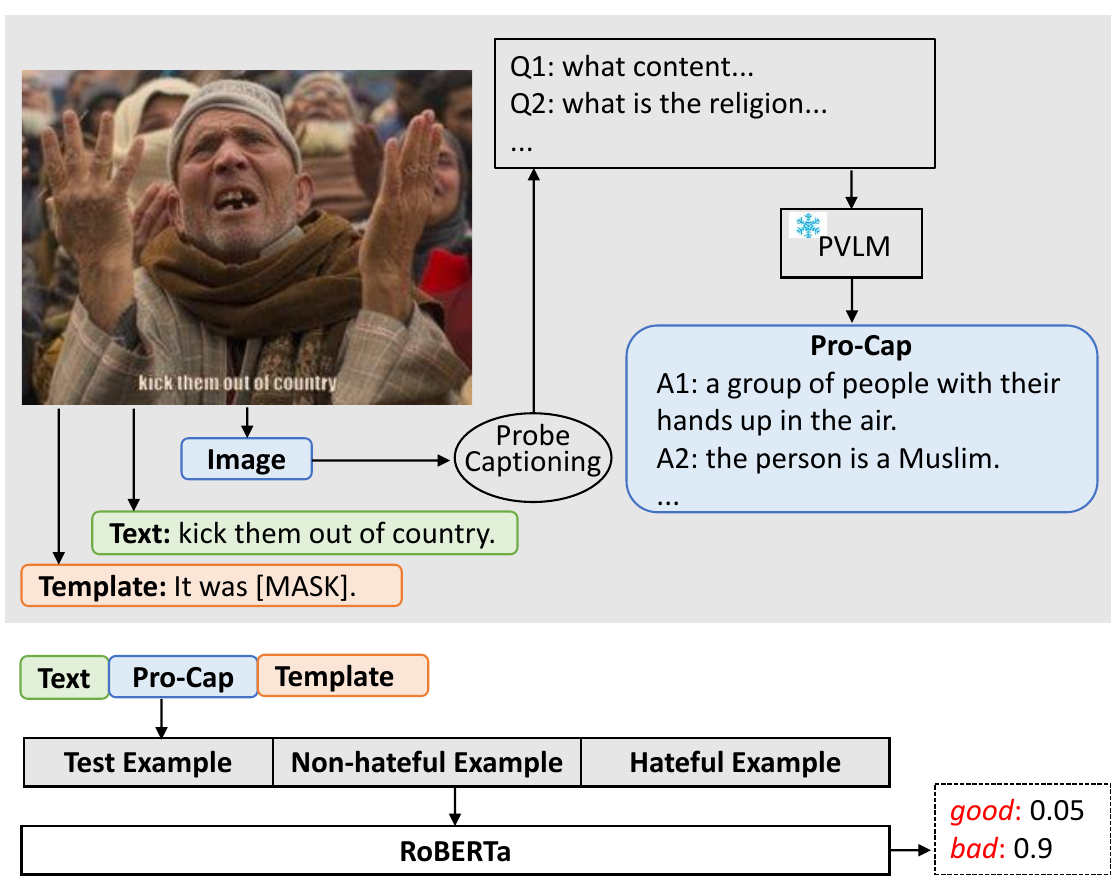} 
 \centering
\caption{An overview of the PromptHate model and how pro-cap is used in PromptHate.} 
	\label{fig:arch}
\end{figure}

\subsection{BERT-based Detection Model}

We now introduce the first of the two alternative hateful meme classification models, which is based on BERT~\cite{devlin2018bert}.
We first feed the concatenation of the meme text $\mathcal{T}$ and the Pro-Cap $\mathcal{C}$ into the BERT model to generate a vector $\mathbf{r} \in \mathbb{R}^d$:
\begin{equation}
    \mathbf{r} = \text{BERT}([\mathcal{T}, \mathcal{C}]),
\end{equation}
where $[\cdot, \cdot]$ represents concatenation. 
Next, we feed the sentence representation $\mathbf{r}$ into a linear layer for hateful meme classification:
\begin{equation}
    \mathbf{s} = \text{Sigmoid}(\mathbf{W}^\text{T}\mathbf{r}+\mathbf{b}),
\end{equation}
where $\mathbf{W} \in \mathbb{R}^{d\times 2}$ and $\mathbf{b}^2$ are learnable parameters. 

\subsection{PromptHate for Hateful Meme Detection}
\label{sec:prompthate}
Next, we introduce the second hateful meme classification model, PromptHate~\cite{DBLP:conf/emnlp/CaoLC022}, which employs a prompt-based method to classify memes. PromptHate was developed to better leverage contextual background knowledge by prompting language models. Given a test meme, PromptHate first uses an image captioning model to obtain generic image captions. It then concatenates the meme text, the image captions, and a prompt template into $\mathcal{S}$: \textit{It was} \texttt{[MASK].}, to prompt a language model (LM) to predict whether the meme is hateful. Specifically, it compares the probability of the language model predicting \texttt{[MASK]} to be a positive word (e.g., \textit{good}) given the context, versus the probability of predicting a negative word (e.g., \textit{bad}). The approach also includes one positive and one negative example in the context, and \texttt{[MASK]} will be replaced by their respective label words. An overview of PromptHate is shown in Figure~\ref{fig:arch}. For further details, please refer to~\cite{DBLP:conf/emnlp/CaoLC022}.

In ~\cite{DBLP:conf/emnlp/CaoLC022}, PromptHate utilizes ClipCap~\cite{DBLP:journals/corr/abs-2111-09734} to generate image captions.
In this work, we replace this with Pro-Cap $\mathcal{C}$. 
We then represent every meme $\mathcal{O}$ as $\mathcal{O}=[\mathcal{T}, \mathcal{C}, \mathcal{S}]$. 
With these inputs, the language models (LMs), for instance, RoBERTa~\cite{DBLP:journals/corr/abs-1907-11692}, generate confidence scores for the masked word over their vocabulary space, $\mathcal{V}$:
\begin{equation}
    \mathbf{p} = \text{Sigmoid}(\text{LM} ([\mathcal{O}_{\text{test}},\mathcal{O}_{\text{non-hate}},\mathcal{O}_{\text{hate}}])),
\end{equation}
where $\mathbf{p} \in \mathbb{R}^{|\mathcal{V}|}$. We extract the score for the label words as the prediction:
\begin{align}
    s_0 &= p_i,\ \mathcal{V}_i=\mathcal{W}_{\text{pos}},\\
    s_1 &= p_j,\ \mathcal{V}_j=\mathcal{W}_{\text{neg}}.
\end{align}

\subsection{Model Training and Prediction}
We denote the ground-truth label of a meme as $\mathbf{\hat{y}} \in \mathbb{R}^2$. If the meme is annotated as \textit{non-hateful}, $\hat{y}_0$ will be $1$ while $\hat{y}_1$ will be $0$, otherwise, $\hat{y}= [0,1]$.
The binary cross-entropy loss is applied for model training:
\begin{equation}
    \text{Loss} = - (\hat{y}_0 * log(s_0) + \hat{y}_1 * log(s_1)).
\end{equation}

For model prediction, if $s_0 > s_1$, the meme will be predicted as non-hateful, otherwise, hateful.


\section{Experiment}
\label{sec:experiment}

In this section, we first introduce our evaluation datasets,  metrics and implementation details. 
Next, we introduce the baselines for comparison. 
Finally, we conduct qualitative analysis with case studies and error analysis to better understand the advantages and limitations of our method.

\subsection{Experiment Settings}

\noindent\textbf{Evaluation Datasets.}
We test our proposed method on benchmarks for hateful meme detection. We evaluate our method on three datasets to better illustrate the generalization and stability of our approach. Table~\ref{tab:dataset} presents the statistics of these datasets.


The \textit{Facebook Hateful Meme} dataset (\textbf{FHM})~\cite{DBLP:conf/nips/KielaFMGSRT20} was constructed by Facebook. 
It contains synthetic memes with added confounders such that unimodal information is insufficient for detection and deep multimodal reasoning is required. 
The FHM dataset contains hateful memes targeting various vulnerable groups in categories including \textit{Religion}, \textit{Race}, \textit{Gender}, \textit{Nationality}, and \textit{Disability}. 
As the labels of the test split of FHM are not available, we performs evaluation on its \textit{dev-seen} split. 

Different from FHM, the \textit{Multimedia Automatic Misogyny Identification}~(\textbf{MAMI}) dataset focuses on a particular type of hateful memes, namely, those targeting women. 
Performance on MAMI therefore reflects the capability of hateful meme detection methods for female victims.

To test our method's generalization capability, we also consider a harmful meme detection dataset, \textbf{HarM}~\cite{DBLP:conf/acl/PramanickDMSANC21}. 
HarM contains memes related to COVID-19, which are classified into three categories: \textit{harmless}, \textit{partially harmful}, and \textit{very harmful}. 
We merge \textit{partially harmful} and \textit{harmful} into one category. 
Because hateful content is always regarded as harmful, we use this dataset to test the capability of generalization of our proposed method from hateful meme detection to harmful meme detection.

\begin{table}[t]
\centering
\caption{Statistical distributions of datasets used for evaluation.}
  \label{tab:dataset}
  \begin{tabular}{c|cc|cc}
    \hline
    \textbf{Datasets}& \multicolumn{2}{c|}{\textbf{Train}} & \multicolumn{2}{c}{\textbf{Test}}\\
    & \#Hate. & \#Non-hate. & \#Hate. & \#Non-hate.\\
    \hline\hline
    FHM & 3,050 & 5,450 &  250 & 250 \\
    HarM & 1,064 & 1,949 & 124 & 230\\
    MAMI & 5,000&5,000 &500 &500 \\
    \hline
\end{tabular}
\end{table}

\noindent\textbf{Evaluation Metrics.}
Hateful meme detection is a binary classification task. 
In addition to detection accuracy, we also compute the Area Under the Receiver Operating Characteristics curve~(AUCROC)  used in prior work~\cite{lippe2020multimodal,zhu2020enhance,DBLP:conf/mm/LeeCFJC21,DBLP:conf/emnlp/CaoLC022}.
We conduct experiments with \textbf{ten} random seeds and report the average performance and standard deviation. 
All models use the same set of random seeds.

\noindent\textbf{Implementation Details.}
Given a meme image, we first detect the meme text with the open-source Easy-OCR tool~\footnote{https://github.com/JaidedAI/EasyOCR} and then in-paint over the detected texts. 
To generate the answers to VQA questions, we prompt BLIP-2~\cite{DBLP:journals/corr/abs-2301-12597}, specifically the FlanT5$_{\text{XL}}$ version. 
We then insert the generated image captions into two text-based hateful meme detection models, i.e., the BERT-based model and the PromptHate model.
For the BERT-based model, to avoid overfitting, we add a dropout rate of $0.4$ to the classification layer. 
We use a learning rate of $2e-5$ and a batch size of $64$. 
For PromptHate, we train the model with a batch size of $16$ and empirically set the learning rate to $1.3e-5$ on FHM and $1e-5$ on the other two datasets~\cite{DBLP:conf/acl/GaoFC20}. 
We optimize both models with the AdamW optimizer~\cite{loshchilov2018fixing} and implement them in PyTorch. Due to space limit, we provide more details (i.e., computation costs and model sizes) in Appendix~\ref{sec:details-imp}.

\subsection{Baselines}
We compare our method against both unimodal and multimodal models to demonstrate the effectiveness of the proposed method, where we regard models receiving information from one modality (i.e., the meme text or the meme image only) as unimodal models. 
Note that because Pro-Cap already contains image information, even if Pro-Cap is input into a unimodal BERT, the model is not considered to be unimodal. 

For the unimodal models, we consider a text-only and an image-only model. 
For the text-only model, we fine-tune a pre-trained BERT model~\cite{devlin2018bert} based on the meme text only for meme classification, which we represent as \textbf{Text-BERT}. 
For the image-only model, we first extract object-level image features with an off-the-shelf feature extractor, Faster-RCNN~\cite{ren2016faster}, which is trained for object detection. 
We then perform average pooling over object features and feed the resulting vector into a classification layer. 
We use \textbf{Image-Region} to denote the image-only model.

For multimodal models, 
we categorize them into two groups: 1) fine-tuning generic multimodal models that are proposed to conduct different multimodal tasks; 2) models specifically designed for hateful meme detection. 
For the first type of multimodal models, we firstly consider the \textbf{MMBT-Region} model~\cite{DBLP:conf/nips/KielaBFT19}, which is a widely used multimodal baseline in hateful meem detection~\cite{DBLP:conf/nips/KielaFMGSRT20,DBLP:conf/emnlp/CaoLC022,DBLP:conf/emnlp/PramanickSDAN021} and the model has not been pre-trained with multimodal data. Secondly, we consider several multimodal pre-trained models, such as   VisualBERT~\cite{li2019visualbert} pre-trained on MS-COCO~\cite{DBLP:conf/eccv/LinMBHPRDZ14}~(\textbf{VisualBERT COCO}) and ViLBERT pre-trained on Conceptual Captions~\cite{DBLP:conf/acl/SoricutDSG18} (\textbf{ViLBERT CC}). 
Some recently released powerful pre-trained models are also included such as the \textit{Align before Fusion} model~\cite{DBLP:journals/corr/abs-2107-07651}~(\textbf{ALBEF}) and the \textit{Bootstrapping Language-Image Pre-training} model~\cite{DBLP:conf/icml/0001LXH22}~(\textbf{BLIP}).
For the second category of baselines which are designed for the meme detection task, we consider the models listed below.
The \textbf{CLIP-BERT} model~\cite{DBLP:conf/emnlp/PramanickSDAN021} leverages the CLIP model~\cite{DBLP:conf/icml/RadfordKHRGASAM21} to deal with noisy meme images, uses pre-trained BERT~\cite{devlin2018bert} for representing meme text, and fuses them with concatenation. 
The \textbf{MOMENTA} model~\cite{DBLP:conf/emnlp/PramanickSDAN021} designed both local and global multimodal fusion mechanisms to exploit  multimodal interactions for hateful meme detection. 
Note that the MOMENTA model is designed to leverage augmented image tags (the detected image entities). 
\textbf{DisMultiHate}~\cite{DBLP:conf/mm/LeeCFJC21} disentangles target information from memes as targets are essential for identifying hateful content. 
The \textbf{PromptHate} model~\cite{DBLP:conf/emnlp/CaoLC022} is 
what we discussed in Section~\ref{sec:prompthate}.

\begin{table*}[t]
\centering
  \caption{Model comparison \textbf{without} any augmented image tags.}
\label{tab:exp-results-wo}
  \begin{tabular}{c|cc|cc|cc}
    \hline
    \textbf{Dataset} &\multicolumn{2}{c|}{\textbf{FHM}}&\multicolumn{2}{c|}{\textbf{MAMI}}&\multicolumn{2}{c}{\textbf{HarM}}\\
    \textbf{Model} & \textbf{AUC.} & \textbf{Acc.}& \textbf{AUC.} & \textbf{Acc.} & \textbf{AUC.} & \textbf{Acc.}\\
    \hline\hline
    Text BERT & 66.10$_{\pm0.55}$& 57.12$_{\pm0.49}$   & 74.48$_{\pm0.60}$ & 67.37$_{\pm0.57}$  & 81.39$_{\pm0.91}$& 75.68$_{\pm1.59}$\\
    Image-Region & 56.69$_{\pm1.05}$ &52.34$_{\pm1.39}$  & 70.20$_{\pm0.63}$ & 64.18$_{\pm0.81}$ &76.46$_{\pm0.47}$ &73.05$_{\pm1.80}$ \\
    \hline\hline
    VisualBERT COCO & 68.71$_{\pm1.02}$& 61.48$_{\pm1.19}$  &78.71$_{\pm0.59}$ &71.06$_{\pm0.94}$  &80.46$_{\pm1.04}$ &75.31$_{\pm1.44}$ \\
    ViLBERT CC& 73.05$_{\pm0.62}$&64.70$_{\pm1.12}$  &77.71$_{\pm1.20}$ &69.48$_{\pm1.00}$  &84.11$_{\pm0.88}$ &78.70$_{\pm1.17}$  \\
    MMBT-Region  & 72.86$_{\pm0.64}$&65.06$_{\pm1.76}$  & 79.17$_{\pm0.91}$& 70.46$_{\pm0.76}$ & 85.48$_{\pm0.75}$& 79.83$_{\pm2.00}$ \\
    \hline
    CLIP-BERT  & 66.97$_{\pm0.34}$&58.28$_{\pm0.63}$  & 77.66$_{\pm0.64}$& 68.44$_{\pm1.07}$ &82.63$_{\pm3.83}$ &80.48$_{\pm1.95}$  \\
    DisMultiHate & 69.11$_{\pm0.84}$& 62.42$_{\pm0.72}$ &78.21$_{\pm0.61}$ & 70.58$_{\pm1.13}$ & 83.69$_{\pm1.33}$& 78.05$_{\pm0.73}$ \\
    PromptHate & 76.76$_{\pm0.95}$&67.82$_{\pm1.23}$  &76.21$_{\pm1.05}$ &68.08$_{\pm0.58}$  &87.51$_{\pm0.74}$ & 79.38$_{\pm1.72}$ \\
    \hline
    BLIP &76.80$_{\pm2.37}$ &69.20$_{\pm1.84}$  & 80.59$_{\pm0.87}$&71.84$_{\pm1.11}$  &87.09$_{\pm1.46}$ &81.81$_{\pm1.74}$  \\
    ALBEF & 79.40$_{\pm0.53}$&70.58$_{\pm0.50}$  & $\mathbf{83.24}_{\pm0.93}$&  72.77$_{\pm1.00}$&85.49$_{\pm1.23}$ &80.99$_{\pm0.80}$  \\
    \hline\hline
    Pro-Cap$_\text{BERT}$ & 77.50$_{\pm0.58}$&  68.14$_{\pm0.64}$& 79.62$_{\pm0.91}$&71.06$_{\pm0.88}$  &89.04$_{\pm1.00}$ &82.06$_{\pm1.92}$ \\
    Pro-Cap$_\text{PromptHate}$  & $\mathbf{80.87}_{\pm0.66}$&$\mathbf{72.28}_{\pm0.90}$  &82.53$_{\pm0.49}$ &$\mathbf{73.06}_{\pm0.82}$  &$\mathbf{90.25}_{\pm0.54}$ &$\mathbf{83.25}_{\pm1.00}$  \\
    \hline
\end{tabular}
\end{table*}


\begin{table*}[t]
\centering
  \caption{Model comparison \textbf{with augmenting the image entities and demographic information}.}
\label{tab:exp-results-with}
  \begin{tabular}{c|cc|cc|cc}
    \hline
    \textbf{Dataset} &\multicolumn{2}{c|}{\textbf{FHM}}&\multicolumn{2}{c|}{\textbf{MAMI}}&\multicolumn{2}{c}{\textbf{HarM}}\\
    \textbf{Model} & \textbf{AUC.} & \textbf{Acc.}& \textbf{AUC.} & \textbf{Acc.} & \textbf{AUC.} & \textbf{Acc.}\\
    \hline\hline
    VisualBERT COCO &72.56$_{\pm0.80}$ &64.28$_{\pm1.27}$  & 80.84$_{\pm0.67}$&72.86$_{\pm0.71}$  &82.96$_{\pm0.98}$ &78.81$_{\pm0.80}$  \\
    ViLBERT CC &75.72$_{\pm0.91}$ &68.24$_{\pm0.44}$  & 80.33$_{\pm1.01}$&71.75$_{\pm1.14}$  &84.79$_{\pm1.23}$ & 81.39$_{\pm1.62}$  \\
    \hline
    MOMENTA  & 69.17$_{\pm4.71}$ & 61.34$_{\pm4.89}$  &81.68$_{\pm2.80}$ &72.10$_{\pm2.90}$   & 86.32$_{\pm3.83}$& 80.48$_{\pm1.95}$\\
    DisMultiHate &79.89$_{\pm1.71}$ & 71.26$_{\pm1.66}$  &80.08$_{\pm0.55}$ & 71.87$_{\pm0.47}$  & 86.39$_{\pm1.17}$ &81.24$_{\pm1.04}$\\
    PromptHate & 81.45$_{\pm0.74}$& 72.98$_{\pm1.09}$ &79.95$_{\pm0.66}$ & 70.31$_{\pm0.64}$ & 90.96$_{\pm0.62}$&84.47$_{\pm1.75}$ \\
    \hline
    BLIP &76.40$_{\pm1.49}$ & 69.29$_{\pm1.44}$ &80.63$_{\pm1.05}$ & 70.62$_{\pm1.48}$ & 86.88$_{\pm1.15}$&  82.66$_{\pm1.13}$\\
    ALBEF & 80.77$_{\pm0.81}$&71.70$_{\pm0.98}$  &82.45$_{\pm0.85}$ &72.45$_{\pm0.96}$  &86.91$_{\pm0.72}$ &  81.78$_{\pm1.20}$\\
    \hline\hline
    Pro-Cap$_\text{BERT}$ & 79.75$_{\pm1.15}$& 71.28$_{\pm0.91}$ & 81.20$_{\pm0.69}$& 71.80$_{\pm1.42}$ & 89.75$_{\pm1.49}$& 82.71$_{\pm1.60}$\\
    Pro-Cap$_\text{PromptHate}$  &$\mathbf{83.58}_{\pm0.60}$  &$\mathbf{75.10}_{\pm0.97}$   &$\mathbf{83.77}_{\pm0.75}$ &$\mathbf{73.63}_{\pm0.75}$  &$\mathbf{91.03}_{\pm1.51}$  & $\mathbf{85.03}_{\pm1.51}$  \\
    \hline
\end{tabular}
\end{table*}

\begin{table}[t]
  \centering
\caption{Ablation study about the impact from the length of VQA answers.}
\label{tab:ablation-length}
  \begin{tabular}{c|c|c|c}
    \hline
    \textbf{Ans. Length} &\textbf{FHM}&\textbf{MAMI}&\textbf{HarM}\\
    \hline
    No Centric &70.08$_{\pm1.57}$ &72.78$_{\pm0.63}$ &80.11$_{\pm1.14}$ \\
    \hline
    Penalty = 1 &71.94$_{\pm0.97}$ & 73.06$_{\pm0.82}$  &82.09$_{\pm1.21}$ \\
    Penalty = 2 &72.28$_{\pm0.90}$ &72.91$_{\pm1.16}$&82.85$_{\pm1.51}$ \\
    Penalty = 3 &71.40$_{\pm1.06}$ &72.47$_{\pm0.74}$ &83.25$_{\pm1.00}$ \\
    \hline
    Pro-Cap$_\text{PromptHate}$ &$\mathbf{72.28_{\pm0.90}}$ &$\mathbf{73.06_{\pm0.82}}$  &$\mathbf{83.25_{\pm1.00}}$ \\
    \hline
\end{tabular}
\end{table}

\begin{table*}[!ht]
\centering
  \caption{Comparison between Pro-Cap$_\text{PromptHate}$ with basic PromptHate. The image caption used by basic PromptHate is denoted as basic caption. Incorrect prediction in {\color{red} red}. The content in $(\cdot)$ of the ground-truth is the target of the hateful meme.}
  \label{tab:case-compare}
  \begin{tabular}{|c|p{4.6cm}|p{4.1cm}|p{4.1cm}| }
    \hline
    \textbf{Meme} & \begin{minipage}[b]{0.5\columnwidth}
		\centering
		\raisebox{-.5\height}{\includegraphics[width=\linewidth]{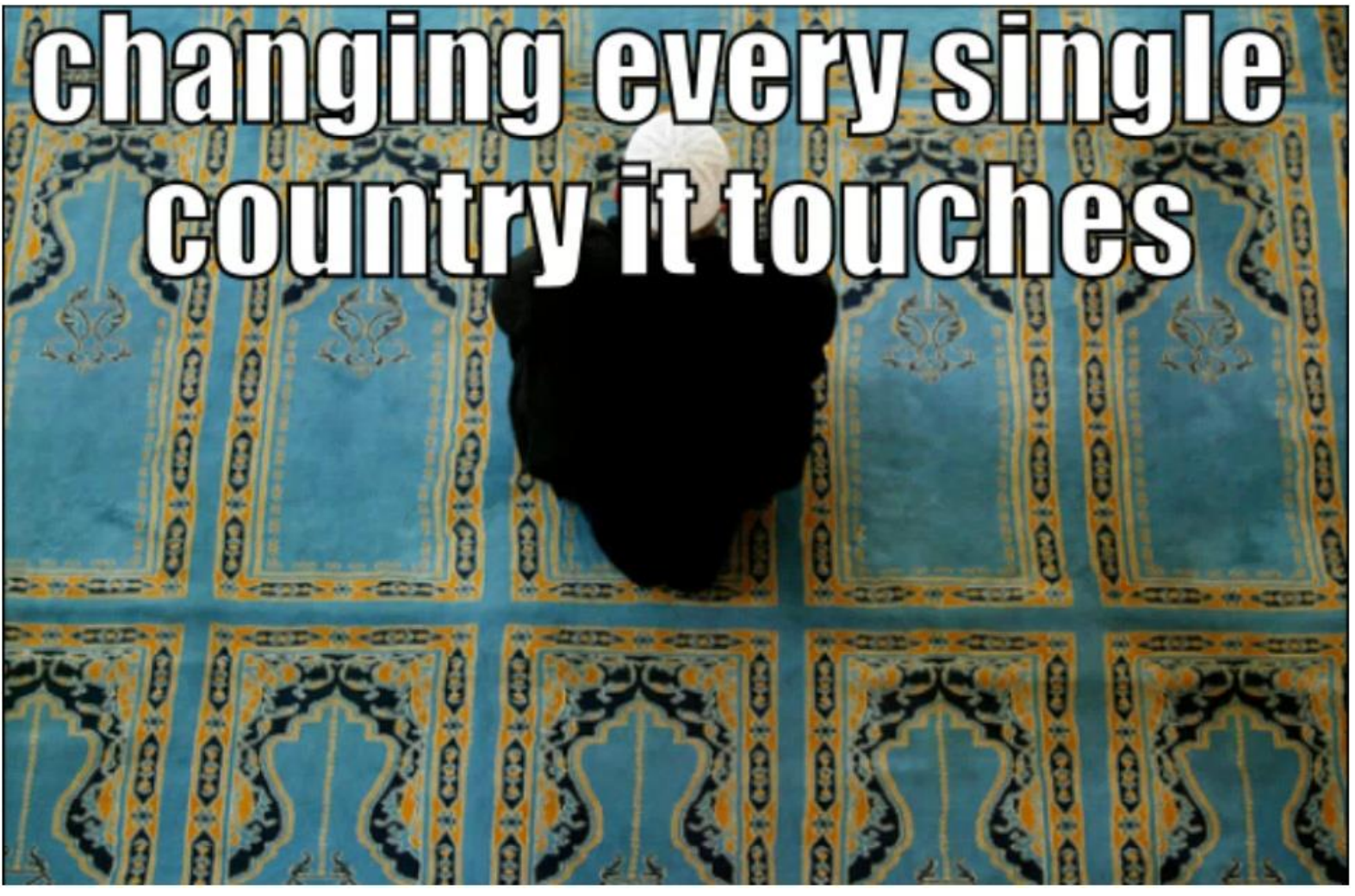}}
	\end{minipage} &
    \begin{minipage}[b]{0.5\columnwidth}
		\centering
		\raisebox{-.5\height}{\includegraphics[width=\linewidth]{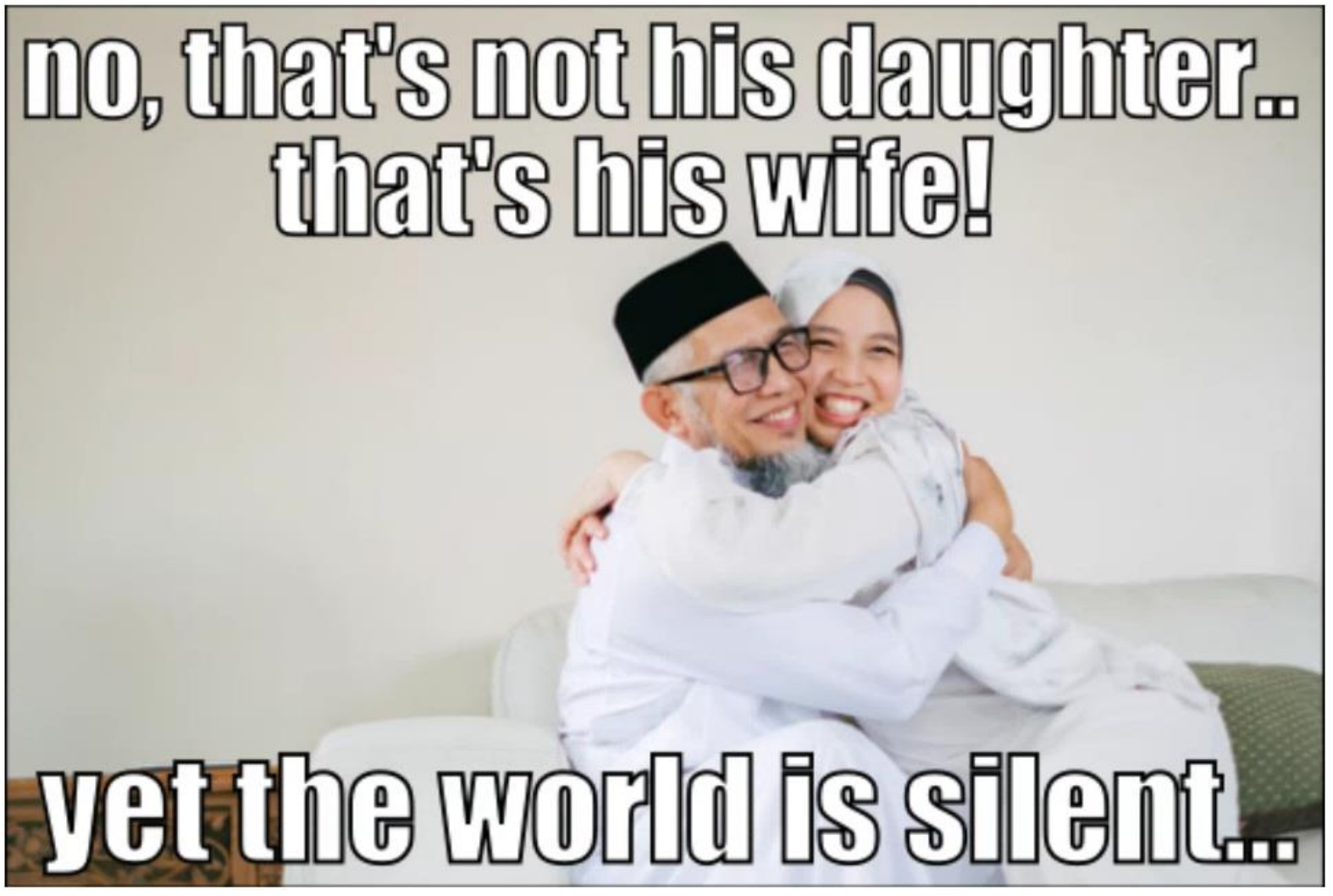}}
	\end{minipage} &
    \begin{minipage}[b]{0.5\columnwidth}
		\centering
		\raisebox{-.5\height}{\includegraphics[width=\linewidth]{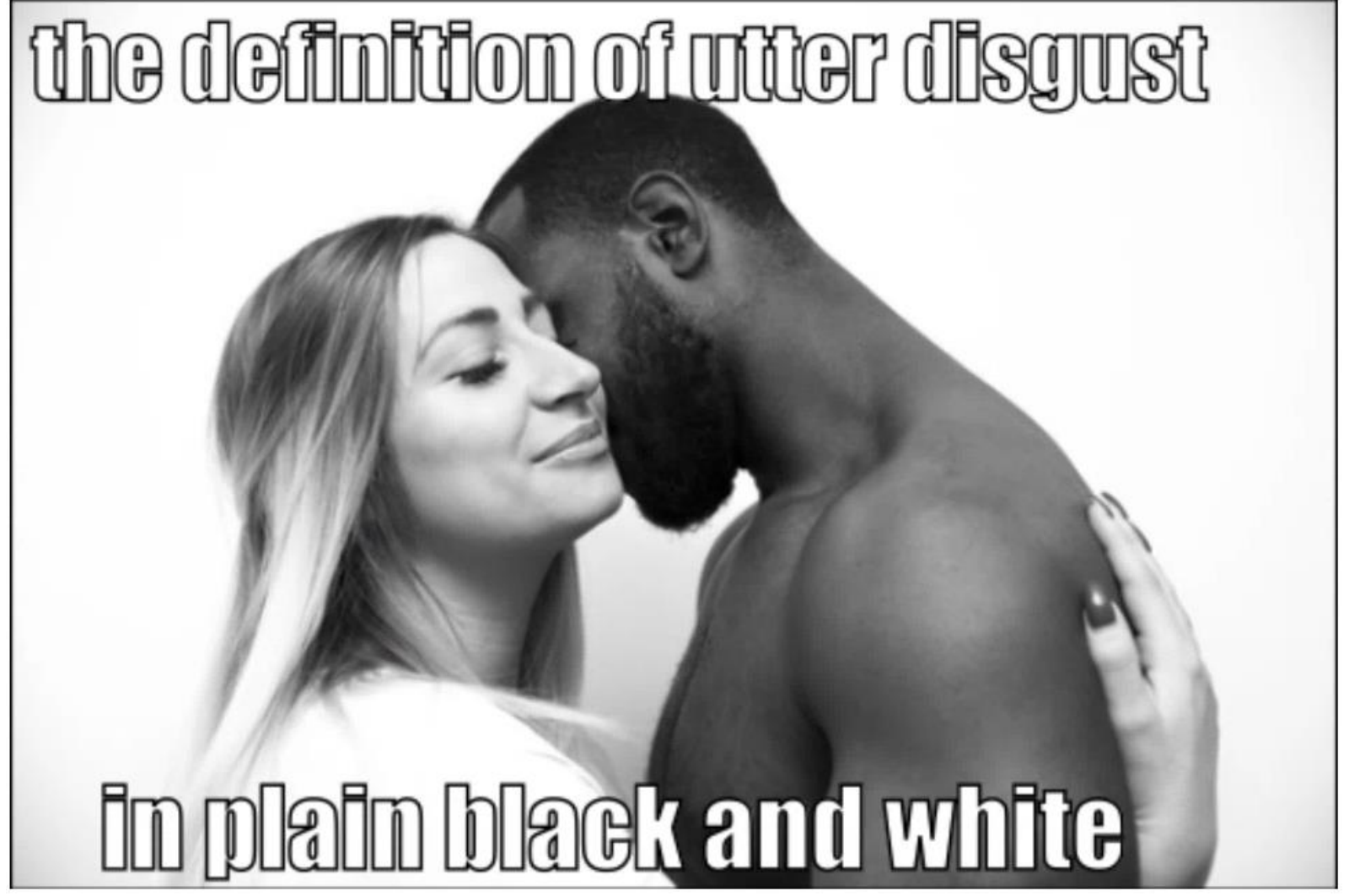}}
	\end{minipage}\\\hline
    \textbf{Ground Truth}  & Hateful (religion) & Hateful (religion) & Hateful (race)\\\hline
    \textbf{Basic PromptHate} & \color{red} Non-hateful & \color{red} Non-hateful & \color{red} Non-hateful\\\hline
    \textbf{Pro-Cap$_\text{PromptHate}$} &Hateful & Hateful & Hateful\\\hline
    \textbf{Meme text} &changing every single country it touches 
    &no that is not his daughter that is his wife yet the world is silent 
    &the definition of utter disgust in plain black and white\\\hline
    \textbf{Basic caption} &mughal structure is one of the largest mosques in the world. 
    & portrait of a father hugging his daughter while smiling at camera in the living room at home.
    & love is in the air!.\\ \hline
    \textbf{Pro-Cap} & (Content:) a black cat sitting on a blue and white tiled floor. (Race:) a black person is standing on a blue and white tiled floor in islamic.
	(Gender:) a man in a black shirt is standing on a blue and white tiled floor with a clock on top of his head.
	(Country:) islamic.
	(Religion:) the person is a muslim and he is wearing a black t-shirt and a black sleeveless.
    & (Content:) a man and a woman hugging on a couch.
	(Race:) a white man and a white woman hugging on a white couch.
	(Gender:) a man and a woman hugging on a white couch.
	(Country:) islamic.
	(Religion:) an muslim man and woman hugging on a white couch.
    &  (Content:) a black and white photo of a man and a woman.
	(Race:) a black man and a white woman in a black and white photo.
	(Gender:) a man and a woman in a black and white photo.
	(Country:) afghanistan.
	(Religion:) he is a christian.\\ \hline
    \end{tabular}
    
\end{table*}

\begin{table}[!ht]
\centering
  \caption{Error cases of Pro-Cap$_\text{PromptHate}$.}
  \label{tab:error}
  \begin{tabular}{|p{0.7cm}|p{3.1cm}|p{3.1cm}| }
    \hline
    \textbf{Meme} & \begin{minipage}[b]{0.38\columnwidth}
		\centering
		\raisebox{-.5\height}{\includegraphics[width=\linewidth]{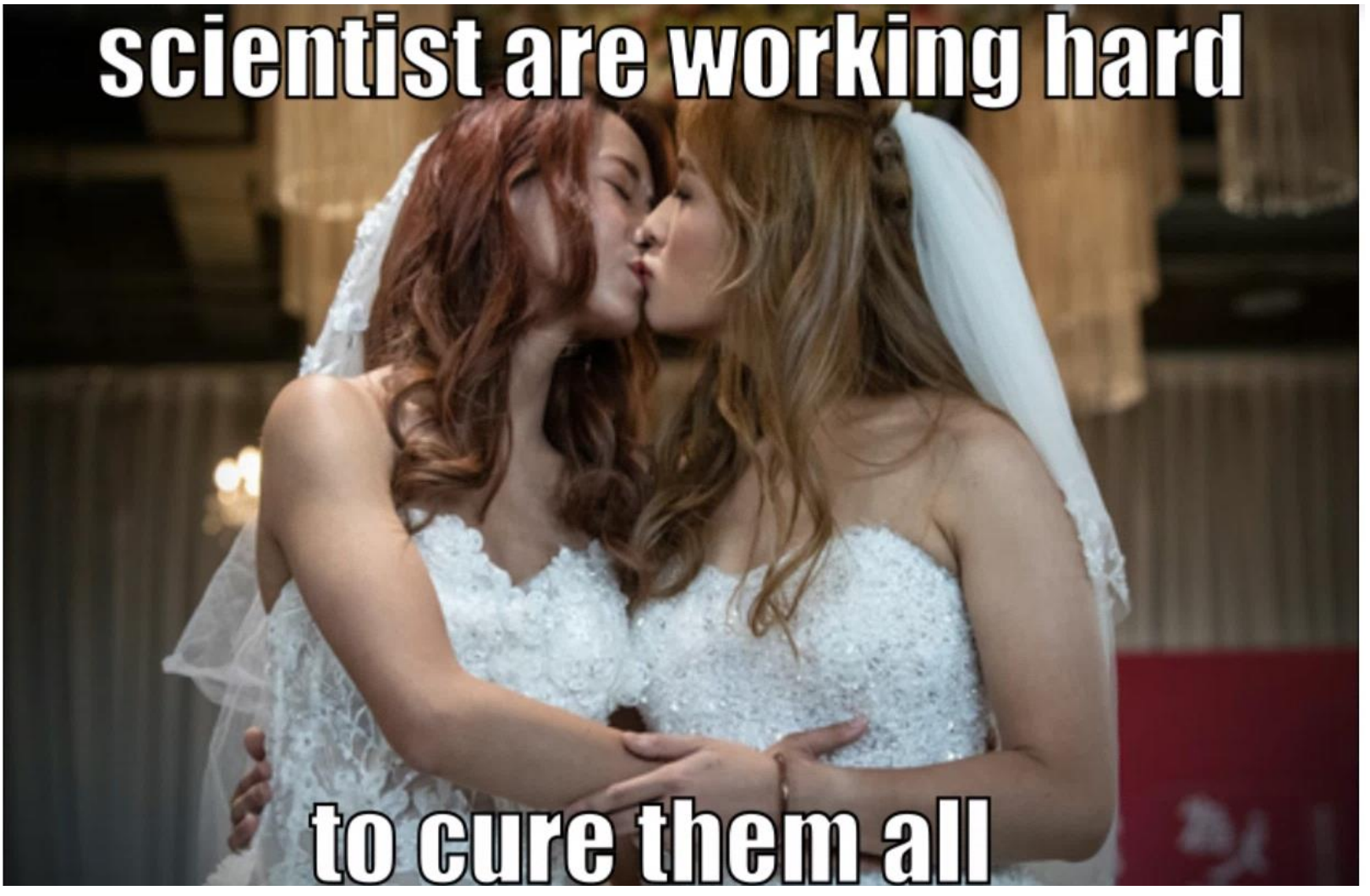}}
	\end{minipage} &
    \begin{minipage}[b]{0.38\columnwidth}
		\centering
		\raisebox{-.5\height}{\includegraphics[width=\linewidth]{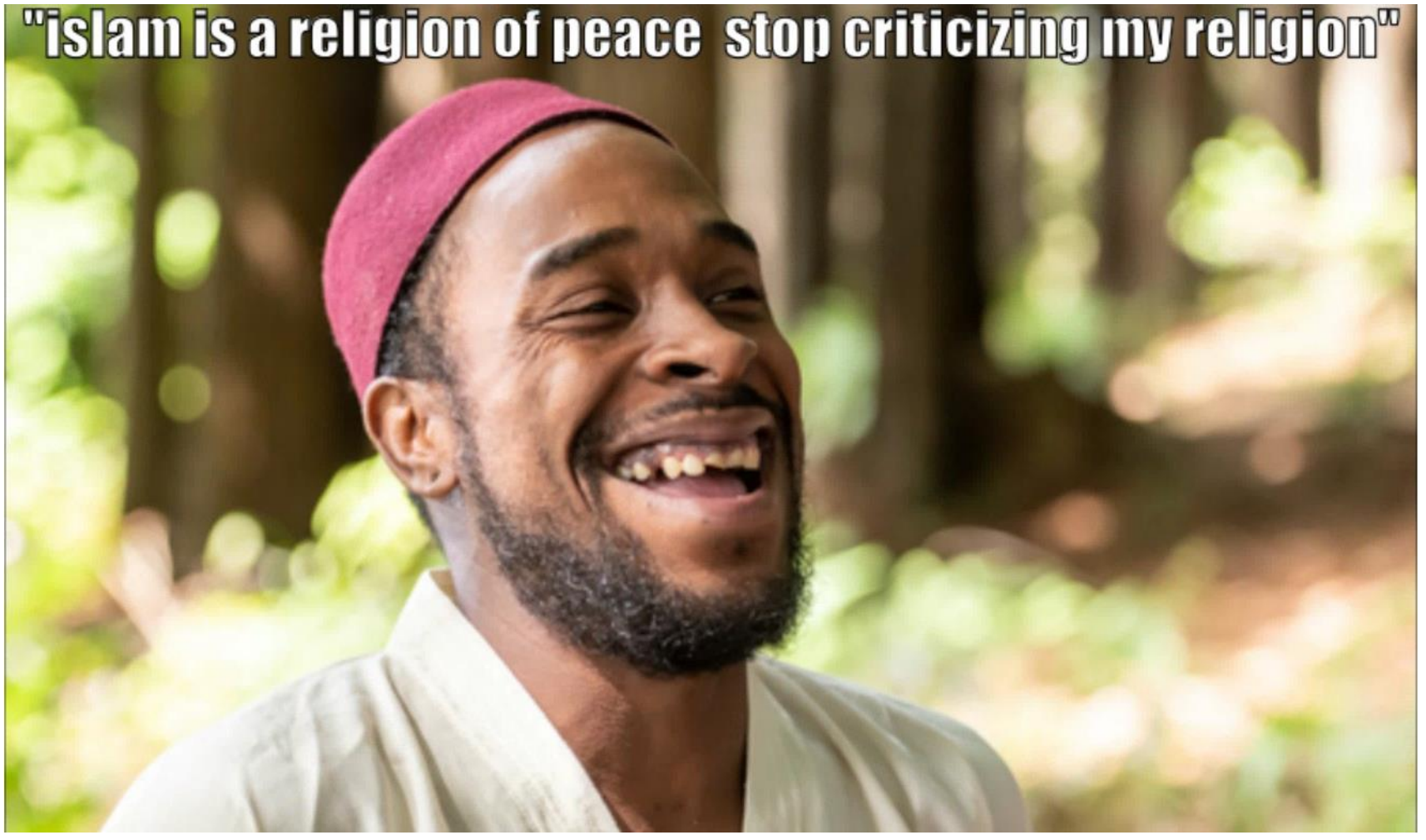}}
	\end{minipage}\\\hline
    \textbf{GT}  & Hateful (gender) & Non-hateful \\\hline
    \textbf{Pred} & \color{red} Non-hateful & \color{red} Hateful \\\hline
    \textbf{Meme text} &scientist are working hard to cure them all 
    &islam is a religion of peace stop criticizing my religion \\\hline
    \textbf{Pro-Cap} &
    (Content:) two women in wedding dresses kissing each other.
	(Race:) a white woman kissing a brunette woman in a wedding dress.
	(Gender:) a woman is kissing a man in a wedding dress.
	(Country:) the person in the image comes from a country in the philippines.
	(Religion:) the person in the image is a christian.
    & (Content:) a man with a beard laughing in the woods.
	(Race:) a african man with a beard and a red hat is smiling in the woods.
	(Gender:) a man with a beard and a red hat in front of a wooded area.
	(Country:) egypt is the country that the person in the image comes from.
	(Religion:) he is a muslim man with a beard and a red tiara on his head.\\ \hline
    \end{tabular}
\end{table}

\subsection{Experiment Results}

As discussed earlier, previous work has shown that additional image tags can enhance hateful meme detection. We therefore consider two settings for comparison: 1) without any augmented image tags; 2) with augmented image tags. We display the performance of models \textbf{without} augmented image tags in Table~\ref{tab:exp-results-wo} and \textbf{with} augmented image tags in Table~\ref{tab:exp-results-with}. The standard deviations (\(\pm\)) of ten random seed runs are also reported, and the best results are highlighted in bold.

\noindent\textbf{Without augmented image tags:} 
We first compare Pro-Cap\(_\text{BERT}\) with unimodal and multimodal models that also utilize BERT as the text encoder (i.e., VisualBERT, ViLBERT, and MMBT-Region). Evidently, Text BERT, which utilizes only meme text, is substantially outperformed by Pro-Cap\(_\text{BERT}\). This suggests that 1) visual signals are vital for hateful meme detection, and 2) the image captions obtained from the probing questions are informative.

Experiment results from multimodal pre-trained BERT-based models are presented in the second block of Table~\ref{tab:exp-results-wo}. 
Interestingly, Pro-Cap$_\text{BERT}$  still has better performances in all three datasets, surpassing the most powerful multimodal pre-trained BERT-base model, ViLBERT, by over $\mathbf{4}\%$ on FHM and surpassing MMBT-Region by about $\mathbf{3}\%$ on HarM. 
This is despite the fact that BERT has less model parameters compared with these multimodal models (e.g, ViLBERT has 252.1M parameters while BERT only has about 110M parameters). 
Pro-Cap$_\text{BERT}$ is still competitive against models specifically designed for hateful meme detection (i.e., models in the third block of Table~\ref{tab:exp-results-wo}). 
We provide experimental results of recently published multimodal pre-trained models (i.e., BLIP and ALBEF) in the fourth block.
By comparing the simple Pro-Cap$_\text{BERT}$ with these models, we observe that Pro-Cap$_\text{BERT}$ gives comparable results. 
While Pro-Cap$_\text{BERT}$ does not out-perform ALBEF and BLIP all the time, performance is reasonably good given that in terms of trainable parameters, Pro-Cap$_\text{BERT}$ is three times smaller than these two pre-trained models. 
Meanwhile, Pro-Cap$_\text{BERT}$ shows even better results than the two models on HarM. 
Notably, HarM is a real-world dataset which is much noisier than FHM. 
HarM also focuses on a relatively new topic (COVID-19), which may not have been observed a lot by the two pre-trained models. 

When comparing BLIP and ALBEF with PromptHate, which has a similar model size, PromptHate with Pro-Cap demonstrates significant advantages over the two models on three benchmarks, especially on the noisy HarM dataset. 
We conjecture that a possible reason is that multimodal pre-trained models leverage pre-training data that is relatively cleaner, on a smaller scale and primarily comprises of non-memes. 
This leads to some difficulties when confronted with noisy real-world memes.  
In contrast pure language models are pre-trained on larger and noisier data, which may lead to some intrinsic robustness. 
If visual signals are reasonably converted to text, pure textual models can be competitive for multimodal tasks such as hateful meme detection. 

Reinforcing the point of proper visual signal conversion, the enhanced performance  of Pro-Cap$_\text{PromptHate}$ over PromptHate highlights the importance of our probing-based captioning method, which provides essential cues for hateful content detection. 
With probe-based captioning, Pro-Cap$_\text{PromptHate}$ is able to conduct deep multimodal reasoning that require background knowledge (due to the good performance on FHM), is stable towards noisy real-world meme data (according to performance on HarM), and has great generalization in meme detection (according to the good performance on all three benchmarks).

\noindent\textbf{With augmented image tags:} For a fair comparison with recent state-of-the-art models, we consider testing our proposed probe-captioning method with the same set of augmented image tags from baselines. To utilize the augmented image tags, we simply pad these tags at the end of each textual meme representation in a similar manner to~\cite{DBLP:conf/emnlp/CaoLC022}. 
With additional image information such as entities and demographic information, most models have some improvements. 
An interesting thing is that neither BLIP nor ALBEF benefits much from additional image tags.
This is because the additional tags are usually single words or short phrases, which may be noisy or redundant, while BLIP and ALBEF may be less capable of dealing with noisy inputs. 
Similar to the results in Table~\ref{tab:exp-results-wo}, when augmenting image information: 1) the simple Pro-Cap$_\text{BERT}$ still obviously surpasses multimodal pre-trained BERT-base models such as VisualBERT or ViLBERT; 2) the Pro-Cap$_\text{BERT}$ performs better than models with similar sizes but specifically designed for hateful meme detection (i.e., MOMENTA or DisMultiHate) in most cases; 3) the Pro-Cap$_\text{BERT}$ achieves comparable results compared with more powerful multimodal pre-trained models, which is about three times larger and surpasses them on the HarM dataset, which is real-world and noisy; 4) Pro-Cap$_\text{PromptHate}$ surpasses the original PromptHate and achieves the best performance on three benchmarks as well.
An interesting point is that comparing Pro-Cap$_\text{PromptHate}$ without any augmented tags and original PromptHate with augmented additional image information, they achieve comparable performance on FHM and HarM and the former even surpasses the latter on MAMI. However, extracting the additional image information is expensive and laborious, which can be replaced by probing-based captioning according to the experimental results.
The equally good performance on three benchmarks highlights the stability and generalization of our proposed approach.


\subsection{Ablation Study}
In this section, we conduct ablation studies to better understand our Pro-Cap method. 
Specifically, we consider the impact of asking different questions and  the impact of the length of answers to the probing questions.
To eliminate other factors, we consider Pro-Cap$_\text{PromptHate}$ without any augmented image tags.
For brevity, we only show accuracy in this section. 
We present the full results in Appendix~\ref{sec:full-ablation}.

\noindent\textbf{The impact of asking hateful-content centric questions: } 
We first conduct an ablation study on the effect of prompting PVLMs with questions facilitating hateful meme detection. 
According to Table~\ref{tab:prompt-ques}, the first question asks about the image content while all questions in the second block are for common vulnerable targets of hateful contents. 
To better understand the impact of including image captions generated by these target-specific questions, we experiment with a setting where captions from the target-specific questions are removed and only the generic caption about image content is used. 
The results are shown in the first block of Table~\ref{tab:ablation-length}. 
Compared with the last block of the table, we observe that with captions generated by target-specific probing questions, the model's performance improved on all three datasets, specifically with over $\mathbf{2}\%$ on FHM and over $\mathbf{3}\%$ on HarM. 
However, we notice minor improvement on MAMI. 
We believe that this is because MAMI memes are all related to woman and generic captions about meme images may already cover the gender of persons in the image. 
However, the other two datasets involve memes with more complexities and therefore asking a wide ragen of target-specific probing questions is more helpful. 
It also implies that in real-world hateful meme detection, probing-based captioning would be helpful. 

\noindent\textbf{The length of answers to probing questions:} 
We apply BLIP-2 as a zero-shot VQA model. 
Different from existing VQA benchmarks~\cite{DBLP:conf/cvpr/HudsonM19,DBLP:conf/cvpr/GoyalKSBP17}, where answers are often single words or short phrases, we may want the answers used as image captions to be longer and thus more informative. 
In this cases, we experiment with answers of different length. 
To conduct the analysis, we set the length penalty in BLIP-2's text decoder for answer generation with different values (i.e., 1, 2 and 3). 
With increased length penalty, longer answers are encouraged. 
We show results of model performance with different answer length in Table~\ref{tab:ablation-length}. 
The results show that detection performance is robust and does not vary much with different answer lengths. 
This indicates the stability of the Pro-Cap method. 
On the other hand, to a very small extent, different datasets do favor answers of different lengths. 
For instance, the HarM dataset prefers longer answers while the MAMI dataset prefers shorter answers.

\subsection{Case Study}
\label{sec:case-vis}
In this section, we conduct case studies to better understand the strengths and limitations of our proposed method. We first compare Pro-Cap$_\text{PromptHate}$ against PromptHate with image captions and show examples in Table~\ref{tab:case-compare}. From the three examples, we observe that in most cases, generic captions about the image content do not provide the key information for hateful meme detection, while asking questions about common vulnerable targets helps. For instance, in the first example, the answer from asking questions about race, country and religion all provide some key words such as \textit{islamic} or \textit{muslim}; in the second example, answers to questions about country and religion are important image captions and the answer to the race-related question is the most important for hateful meme detection. In contrast, we observe that the basic captions in the original PromptHate miss these crucial facts about the meme images.

Next, we conduct error analysis about our proposed probe-captioning in Table~\ref{tab:error}. In the first example, all probe-captions generate sufficient image captions for the hateful meme detection, while the model still fails at prediction. This may be due to the current language models performing poorly in further complex reasoning. We also note that the small scale of hateful meme  datasets may be inadequate for training a model to perform complex reasoning. Recent studies about large language models pre-trained with trillions of words~\cite{DBLP:journals/corr/abs-2302-13971} may facilitate hateful meme detection to some extent. Besides, we observe minor errors in predicted answers from the zero-shot VQA model (e.g., the wrong prediction of ``a woman kissing a man'' when asking about gender). It highlights that with the development of better zero-shot VQA models, the our strategy could potentially facilitate more for the two text-based hateful meme detection models.
The second example highlights a limitation of most hateful content detection models in that they may be biased. During the training stage, there may be hateful contents towards Muslims so that once models seen Muslims, they tend to predict the meme as hateful. To alleviate the issue, debiasing techniques may be needed. Due to space limitation, we omit visualization examples in the main pages and refer the reader to examples in Appendix~\ref{sec:vis-cases}.

\section{Conclusion}

In this study, we attempt to leverage pre-trained vision-language models (PVLMs) in a low-computation-cost manner to aid the task of hateful meme detection. Specifically, without any fine-tuning of PVLMs, we probe them in a zero-shot VQA manner to generate hateful content-centric image captions. With the distilled knowledge from large PVLMs, we observe that a simple language model, BERT, can surpass all multimodal pre-trained BERT models of a similar scale. PromptHate with probe-captioning outperforms previous results significantly and achieves the new state-of-the-art on three benchmarks.

\noindent\textbf{Limitations: }We would like to point out a few limitations of the proposed method, suggesting potential future directions. Firstly, we heuristically use answers to all probing questions as Pro-Cap, even though some questions may be irrelevant to the meme target. We report the performance of \textsf{PromptHate} with the answer from one probing question in Appendix~\ref{sec:appendix-individual}, highlighting that using all questions may not be the optimal solution. A future direction could involve training a model to dynamically select probing questions that are most relevant for meme detection. Secondly, although we demonstrate the effectiveness of Pro-Cap through performance and a case study in this paper, more thorough analysis is needed. For instance, in the future, we could use a gradient-based interpretation approach~\cite{DBLP:conf/iccv/SelvarajuCDVPB17} to examine how different probing questions influence the final results, thereby enhancing the interpretation of the models.


\clearpage
\bibliographystyle{ACM-Reference-Format}
\balance
\bibliography{ref}

\clearpage
\section*{APPENDIX}
\appendix
\label{sec:appendix}

\begin{table}[ht]
  \centering
  \caption{Comparison between Pro-Cap$_\text{PromptHate}$ and basic PromptHate on HarM dataset.}
  \label{tab:vis-comp-harm}
  \begin{tabular}{|p{1cm}|p{3.5cm}|p{3.5cm}| }
    \hline
    \textbf{Meme} & \begin{minipage}[b]{0.4\columnwidth}
		\centering
		\raisebox{-.5\height}{\includegraphics[width=\linewidth]{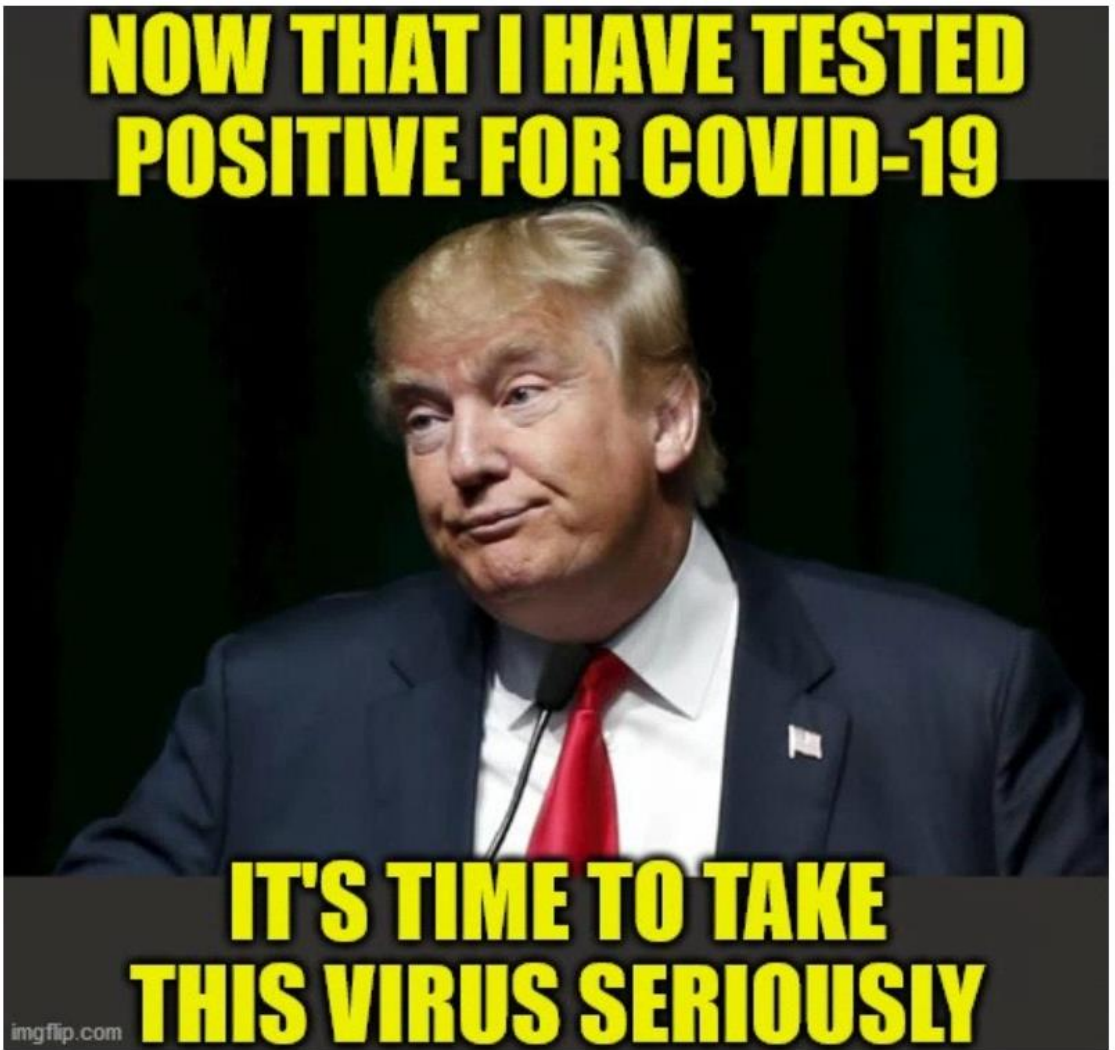}}
	\end{minipage} &
    \begin{minipage}[b]{0.4\columnwidth}
		\centering
		\raisebox{-.5\height}{\includegraphics[width=\linewidth]{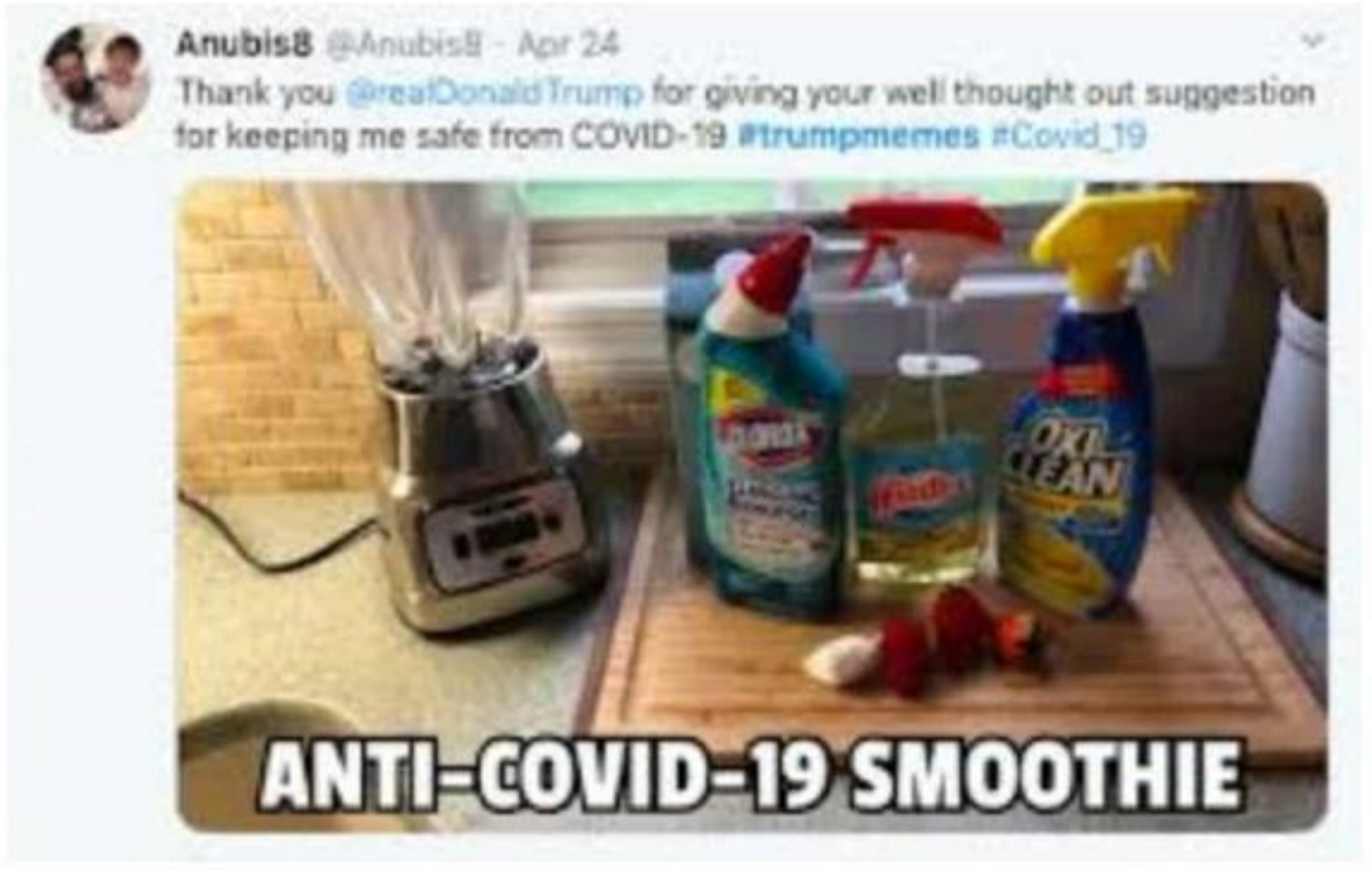}}
	\end{minipage}\\\hline
    \textbf{Ground Truth}  & Hateful & Hateful \\\hline
    \textbf{Basic Pred} & \color{red} Non-hateful & \color{red} Non-hateful \\ \hline
    \textbf{Pro Pred}  &Hateful &Hateful\\ \hline
    \textbf{Meme text} & now that I have tested positive for COVID-19. It's time to take this virus seriously.
    & Thank you reatDonald Trump for giving your well thought out suggestionfor keeping me safe from COVID-19 atrumpmemes \#Covid 19 Anti-covid-19 smoothie\\\hline
    \textbf{Basic Caption} &i'm going to get tested for a virus!. & how to clean a kitchen sink with vinegar and food coloring.\\ \hline
    \textbf{Pro-Cap} &
    (Generic:) trump in a suit and tie with the caption now that positive covid-19 it's time to take the virus seriously.
    (Race:) he is a white man in a suit and tie with a red tie and a white hat with a red hat.
    (Gender: he is a man in a suit and tie with a caption that says now that positive covid-19 it's time to take.
    (Country):  us of america.
    (Religion: he is a christian.
    & (Generic:) a picture of a blender with cleaning products on it.\\ \hline
    \end{tabular}
    
\end{table}

\begin{table}[ht]
\small
  \centering
  \caption{Comparison between Pro-Cap$_\text{PromptHate}$ and basic PromptHate on MAMI dataset.}
  \label{tab:vis-comp-mimc}
  \begin{tabular}{|p{1cm}|p{3.5cm}|p{3.5cm}| }
    \hline
    \textbf{Meme} & \begin{minipage}[b]{0.4\columnwidth}
		\centering
		\raisebox{-.5\height}{\includegraphics[width=\linewidth]{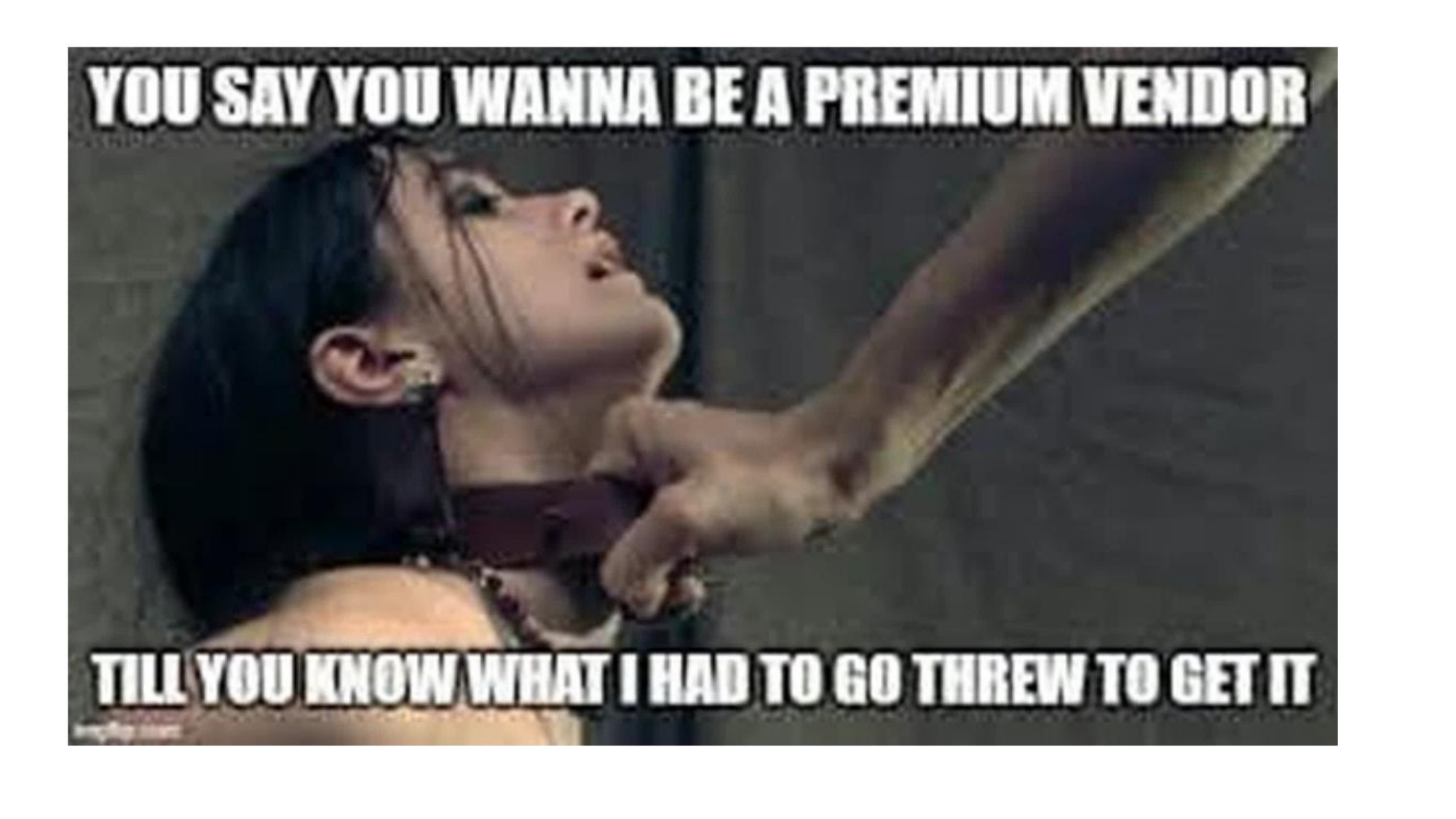}}
	\end{minipage} &
    \begin{minipage}[b]{0.4\columnwidth}
		\centering
		\raisebox{-.5\height}{\includegraphics[width=\linewidth]{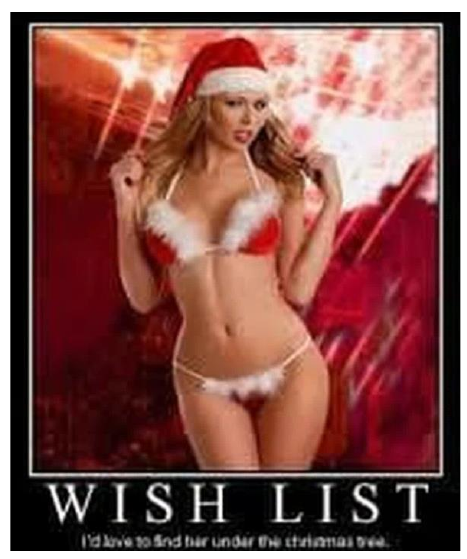}}
	\end{minipage}\\\hline
    \textbf{Ground Truth}  & Hateful & Hateful \\\hline
    \textbf{Basic Pred} & \color{red} Non-hateful & \color{red} Non-hateful \\ \hline
    \textbf{Pro Pred}  &Hateful &Hateful\\ \hline
    \textbf{Meme text} & you say you want to be a premium vendor until you know what I had to go threw to get it.
    & wish list I would love to send her under the Christmas eve.\\\hline
    \textbf{Basic Caption} & wallpaper probably with a well dressed person and a well dressed person entitled person.
    & western christian holiday from all of us!. \\ \hline
    \textbf{Pro-Cap}& 
    (Generic): a woman is being choked by a man.
    (Race): a black woman is being choked by a man in a t-shirt with a picture of a t-shirt.
    (Gender): a woman is being choked by a man in the image.
    (Country): afghanistan.
    (Religion): the person in the image is a christian.
    &(Generic): a woman in a santa claus hat posing in a bikini.
    (Race): a white woman in a santa claus hat posing in a sexy bikini.
    (Gender): a woman wearing a santa claus hat and a bikini.
   (Country): switzerland.
    (Religion): santa claus is the religion of the person in the image. \\ \hline
    \end{tabular}
    
\end{table}

\begin{table}[ht]
\centering
\caption{Number of parameters in VQA models.}
  \label{tab:num-params}
  \begin{tabular}{c|c}
    \toprule
    \textbf{Method} & \textbf{\# Params (M)}  \\
    \midrule
    Text BERT & 109.9\\
    Image Region &  1.0\\ 
    \hline 
    Visual BERT COCO & 111.8 \\
    ViLBERT CC &252.1 \\
    MMBT-Region & 111.5  \\
    \hline
    CLIP BERT  &111.7\\
    MOMENTA &71.9\\
    DisMultiHate &115.6 \\
    \hline
    BLIP & 385.0\\
    ALBEF &209.5 \\
    \hline
    BERT & 109.9 \\
    PromptHate & 355.4\\
    \midrule
\end{tabular}
\end{table}

\begin{table*}[t]
  \small
  \centering
  \begin{tabular}{c|cc|cc|cc}
    \hline
    \textbf{Ans. Length} &\multicolumn{2}{c|}{\textbf{FHM}}&\multicolumn{2}{c|}{\textbf{MAMI}}&\multicolumn{2}{c}{\textbf{HarM}}\\
    \textbf{Model} & \textbf{AUC.} & \textbf{Acc.}& \textbf{AUC.} & \textbf{Acc.} & \textbf{AUC.} & \textbf{Acc.}\\
    \hline
    No Centric & 79.08$_{\pm0.94}$&70.08$_{\pm1.57}$ &82.26$_{\pm0.71}$ &72.78$_{\pm0.63}$ &87.04$_{\pm0.89}$ &80.11$_{\pm1.14}$ \\
    \hline
    Penalty = 1 & 80.76$_{\pm1.06}$&71.94$_{\pm0.97}$ & 82.53$_{\pm0.49}$&73.06$_{\pm0.82}$ &88.34$_{\pm0.77}$ &82.09$_{\pm1.21}$ \\
    Penalty = 2 &80.87$_{\pm0.66}$ &72.28$_{\pm0.90}$ &82.27$_{\pm0.57}$ &72.91$_{\pm1.16}$ &90.25$_{\pm0.72}$ &82.85$_{\pm1.51}$ \\
    Penalty = 3 & 79.62$_{\pm0.93}$& 71.40$_{\pm1.06}$&82.36$_{\pm0.97}$ &72.47$_{\pm0.74}$ &90.25$_{\pm0.54}$ &83.25$_{\pm1.00}$ \\
    \hline
\end{tabular}
\caption{Model comparison \textbf{without} any augmented image tags.}
\label{tab:full-ablation-length}
\end{table*}

\begin{table*}[t]
\centering
  \caption{Model performance when only asking a single probing question.}
\label{tab:exp-results-single-target}
  \begin{tabular}{c|cc|cc|cc}
    \hline
    \textbf{Dataset} &\multicolumn{2}{c|}{\textbf{FHM}}&\multicolumn{2}{c|}{\textbf{MAMI}}&\multicolumn{2}{c}{\textbf{HarM}}\\
    \textbf{Model} & \textbf{AUC.} & \textbf{Acc.}& \textbf{AUC.} & \textbf{Acc.} & \textbf{AUC.} & \textbf{Acc.}\\
    \hline\hline
    Race &$83.63_{\pm0.26}$ &$74.28_{\pm1.34}$ &$84.00_{\pm0.57}$ &$73.51_{\pm1.10} $& $90.43_{\pm0.70}$&$82.26_{\pm1.96}$ \\
    Gender & $83.91_{\pm0.97}$&$76.08_{\pm1.47}$ & $84.34_{\pm1.06}$&$74.21_{\pm0.64}$ & $91.05_{\pm0.57}$&$83.16_{\pm1.79}$ \\
    Religion & $84.85_{\pm0.87}$&$75.52_{\pm1.45}$ &$83.90_{\pm0.78}$ &$73.95_{\pm0.84}$ & $90.86_{\pm0.39}$&$82.15_{\pm1.15}$ \\
    Nationality & $85.78_{\pm0.37}$&$75.72_{\pm0.96}$ &$83.73_{\pm0.49}$ &$72.76_{\pm0.52}$ & $91.27_{\pm0.68}$&$84.30_{\pm1.82}$ \\
    Disability &$85.26_{\pm0.64}$ &$75.96_{\pm0.82}$ & $83.81_{\pm0.87}$&$73.75_{\pm0.76}$ &$90.20_{\pm0.82}$ &$84.12_{\pm0.60}$ \\
    Animal & $84.93_{\pm0.31}$&$75.48_{\pm0.72}$ & $84.10_{\pm0.49}$ &$73.53_{\pm0.90}$ & $90.13_{\pm0.87}$ &$82.65_{\pm2.01}$ \\
    \midrule
    Pro-Cap$_\text{PromptHate}$  &$\mathbf{83.58}_{\pm0.60}$  &$\mathbf{75.10}_{\pm0.97}$   &$\mathbf{83.77}_{\pm0.75}$ &$\mathbf{73.63}_{\pm0.75}$  &$\mathbf{91.03}_{\pm1.51}$  & $\mathbf{85.03}_{\pm1.51}$  \\
    \hline
\end{tabular}
\end{table*}
\section{Details for Implementation}
\label{sec:details-imp}
We implement all models under the PyTorch Library with the CUDA-11.2 version. We use the Tesla V 100 GPU, each with a dedicated memory of $32$GB. For models specifically implemented for hateful meme detection, we take the codes published from the author for re-implementation~\footnote{CLIP-BERT/MOMENTA: https://github.com/LCS2-IIITD/MOMENTA;DisMultiHate: https://gitlab.com/bottle\_shop/safe/dismultihate; PromptHate: https://gitlab.com/bottle\_shop/safe/prompthate}. For pre-trained models which can be found under the Huggingface Library, we use the packages from Huggingface~\footnote{https://huggingface.co/}, specifically the BERT~\cite{devlin2018bert}, VisualBERT~\cite{li2019visualbert} and the BLIP model. Gor ViLBERT~\cite{lu2019vilbert}, we take the released code from the authors~\footnote{https://github.com/facebookresearch/vilbert-multi-task}. For  ALBEF~\cite{DBLP:journals/corr/abs-2107-07651} and BLIP-2~\cite{DBLP:journals/corr/abs-2301-12597}, we use the packages under the LAVIS Library\footnote{https://github.com/salesforce/LAVIS}.

For each meme image, we constrain the total length of the meme text and the generic image caption (either from the captioning model or by asking about the content of the image) to be $65$. For each additional questions, we restrict its length to be shorter than $20$. If the concatenation of the sentence exceeds the limited length, the sentence will be truncated, otherwise, if the sentence is shorted than the limited length, it will be padded. We set the number of training epochs to be $10$ for all models.

The number of model parameters are summarized in Table~\ref{tab:num-params}.

\section{Full Ablation Study Results}
\label{sec:full-ablation}
Due to the limitation of space, we only show results of accuracy in ablation studies in Table~\ref{tab:ablation-length}. The full results including both the AUC and the accuracy are provided in Table~\ref{tab:full-ablation-length}.

\section{Visualization Cases}
\label{sec:vis-cases}
In Section~\ref{sec:case-vis}, we provide visualization of cases for comparing Pro-Cap$_\text{PromptHate}$ with the basic PromptHate. Due to space constraints, we omit examples from the other two datasets. We provide more visualization cases in this part. The cases from the HarM dataset are illustrated in Table~\ref{tab:vis-comp-harm} and the cases from the MAMI dataset are shown in Table~\ref{tab:vis-comp-mimc}.

\section{Results with Pro-Cap about One Target}
\label{sec:appendix-individual}
In Section~\ref{sec:experiment}, we only report results when models use Pro-Cap from all probing questions. In this part, we report results (with entities) when using the answers from a single probing question in Table~\ref{tab:exp-results-single-target}.

According to the results, we observe models using answers to a single probing question are all powerful and some even surpass heuristically asking all probing questions (e.g., using the question asking about \textit{nationality} on FHM is better than using all probing questions). It points out using all probing captions may not be the optimal solution and may generate irrelevant image descriptions. For instance, confronted with a hateful meme targeting at black people, it is meaningless to ask the religion of people in the image. Interestingly, on MAMI, when only using answers to the probing question about gender reaches teh best performance. It is because MAMI contains only hateful memes about woman. A promising direction would train the model to dynamically select probing questions essential for meme detection for different memes.

\end{document}